# The Largest Compatible Subset Problem for Phylogenetic Data

Andy Auyeung[1] and Ajith Abraham[2]

[1] Department of Computer Science, Oklahoma State University,
Stillwater, OK 74078, USA
`wingha@cs.okstate.edu`

[2] Department of Computer Science, Oklahoma State University,
Tulsa. OK 74106, USA
`Ajith.abraham@ieee.org`

**Abstract.** The phylogenetic tree construction is to infer the evolutionary relationship between species from the experimental data. However, the experimental data are often imperfect and conflicting each others. Therefore, it is important to extract the motif from the imperfect data. The largest compatible subset problem is that, given a set of experimental data, we want to discard the minimum such that the remaining is compatible. The largest compatible subset problem can be viewed as the vertex cover problem in the graph theory that has been proven to be NP-hard. In this paper, we propose a hybrid Evolutionary Computing (EC) method for this problem. The proposed method combines the EC approach and the algorithmic approach for special structured graphs. As a result, the complexity of the problem is dramatically reduced. Experiments were performed on randomly generated graphs with different edge densities. The vertex covers produced by the proposed method were then compared to the vertex covers produced by a 2-approximation algorithm. The experimental results showed that the proposed method consistently outperformed a classical 2-approximation algorithm. Furthermore, a significant improvement was found when the graph density was small.

## 1 Introduction

The study of phylogenetic (phylogeny) is to understand the evolutionary relationships between species [17]. Different models have been proposed to model the evolutionary relationships between species from different types of experimental data [11]. The perfect phylogeny is a tree-based model that uses binary characters to infer phylogeny [10]. Due to the noisy nature of the experimental data, conflicts are often found between subsets of the experimental data. Therefore, it is important to extract the motif from the imperfect data. The largest compatible subset problem is that, given a set of experimental data, we want to discard the minimum such that the remaining is compatible [6].

In the case of the perfect phylogeny, the conflicts between subsets of data come in pair-wise form [8]. In other words, the compatibility between subset $A$ and subset $B$ are independent from other subsets. Moreover, if $A$ is incompatible with $B$, $B$ is also incompatible with $A$. From this condition, we model the incompatibility between subsets of data by a graph, where each vertex represents a disjointed data subset, and an edge $(u, v)$ indicates that vertex $u$ and $v$ are incompatible.

In the context of a graph $G=(V, E)$, where $V$ is the set of vertices and $E$ is the set of edges, the largest compatible subset problem is to find a subset $U' \subseteq V$, such that $(u, v) \notin E$, $\forall u, v \in U'$ and $|U'|$ is maximum. Equivalently, we can find a subset $V' \subseteq V$ such that $\forall (u, v) \in E$, then either $u \in V'$ or $v \in V'$ (or both) and $|V'|$ is minimum. In fact, this problem has already been studied in the graph theory and is called the vertex cover problem [15].

The vertex cover problem has been proven to be NP-hard [5]. There is no efficient algorithm to solve it in general. Therefore, many heuristic algorithms have been proposed [14]. Although good approximation algorithms have been proposed for graphs with a bounded edge degree [12], the vertex cover problem does not have a good approximation solution in general.

On the other hand, for special structured graphs, such as simple cycles and trees, there are linear-time algorithms to solve them. Therefore, although efficient solution cannot be found for the whole graph in general, the optimal solution can easily be obtained for some special structured components in the graph.

The Evolutionary Computing (EC) is a powerful searching technique that is widely used in computational biology problems [3]. In this paper, we propose an EC approach for the largest compatible subset problem. The idea of the proposed method is to make use of the efficient algorithms for special structured graphs in the EC approach, so that the search space can be dramatically reduced. Therefore, the proposed method is effective and efficient. Our method does not provide a guaranteed error bound. However, our experiments show that it consistently found better solutions than a classical 2-approximation algorithm.

The rest of the paper is organized as follows. In Section 2, some background materials about phylogenetics and the vertex cover problem are presented. In Section 3, the proposed method is explained. In Section 4, the experimental setup and results are shown. In Section 5, observations from the experiments and some design issues are discussed. Finally in Section 6, some concluding remarks are made.

## 2 Related Research

### 2.1 The Perfect Phylogeny

The perfect phylogeny is a phylogenetic model that uses binary characters. An $m$ by $n$ 0-1 matrix $M$ records the exhibitions of the n characters in the $m$ species. That is $M[i, j] = 1$ if and only if species $i$ exhibits character $j$, and it is equal to zero otherwise. The definition of a perfect phylogenetic tree for matrix $M$ is as follows. A rooted tree $T$ is a perfect phylogenetic tree of $M$ if it satisfies: 1, $T$ has exactly $m$ leaves; 2, each of the $m$ species labels exactly one leaf of $T$; 3, each of the $n$ characters labels exactly one edge of $T$; and 4, for any leaf $l$, the unique path from the root node to $l$ contains labels specify all the characters that $l$ (its corresponding species) has. Figure 1 shows an example of a matrix $M$ and its perfect phylogenetic tree $T$.

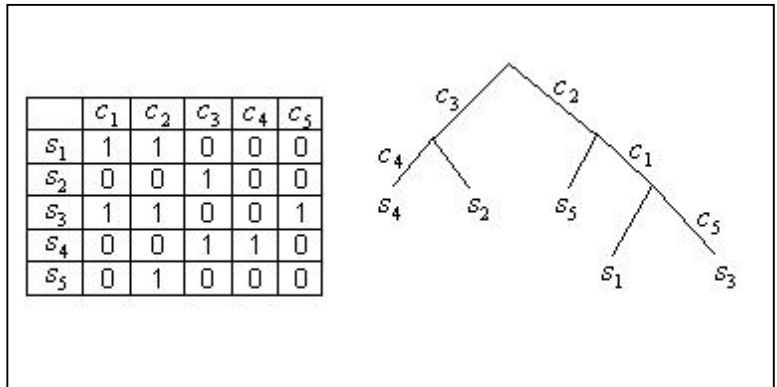

Figure 1. A 5 x 5 0-1 matrix $M$ and its perfect phylogenetic tree $T$.

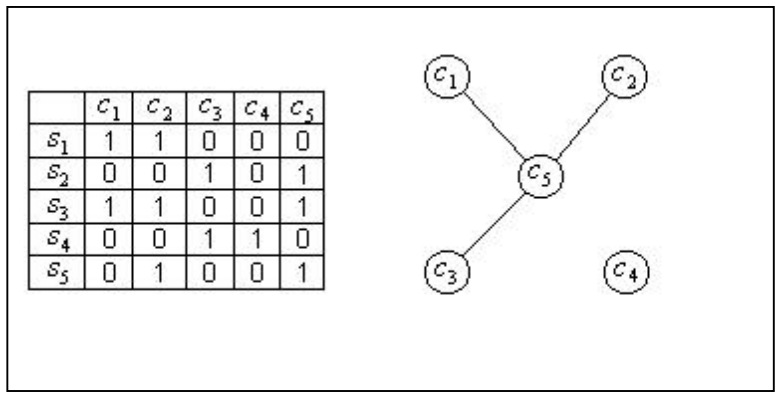

Figure 2. A matrix $M$ that is not a perfect phylogeny and its conflict graph.

Not all matrices are perfect phylogenys. For any two characters $c_i$ and $c_j$, let $O_i$ and $O_j$ be the sets of species who exhibits $c_i$ and $c_j$ respectively. The matrix $M$ is a perfect phylogeny if and only if $O_i$ and $O_j$ are either disjoint or one contains the other, for every $i, j$. In other words, the character $c_i$ and $c_j$ are incompatible when $O_i$ and $O_j$ are overlapped. The incompatibility between the characters in $M$ can be visualized by a conflict graph, where each vertex represents a character and each edge

```
Algorithm A [G = (V, E)]
{   V' = 0
    while (E is not empty)
    {   Pick an arbitrary edge (u, v) ∈ E
        V' = V' ∪ {u, v}
        Remove (u, v) and all edges incident
        on u or v from E
    }
    return V'
}
```

Algorithm A. A 2-approximation algorithm that finds a vertex cover.

```
Algorithm B [G=(V, E)]
{   V' = 0
    while (E is not empty)
    {
        Pick an arbitrary leaf node u in G
        Let v be the parent node of u
        V' = V' ∪ {v}
        Removing v and all edges incident on
        v from G
    }
    return V'
}
```

Algorithm B. A linear-time algorithm that finds the optimal vertex cover on tree.

represents a conflict. Figure 2 shows an example of a matrix $M$ that is not a perfect phylogeny and its conflict graph.

In practice, most experimental data are not perfect phylogenys. Therefore, we want to extract a matrix $M'$ from $M$, such that $M'$ is a perfect phylogeny. The largest compatible subset problem for the perfect phylogeny is that, given an $m$ by $n$ matrix $M$, we want to find an $m$ by $n'$ matrix $M'$ by removing columns from $M$, such that $n'$ is maximized and $M'$ is a perfect phylogeny. In the context of the conflict graph, we want to find the largest subset of vertices $U'$ to keep, such that the resulted graph is edge free.

```
Algorithm C [G=(V, E)]
{   V' = V
    for each critical vertex u
    {   if u has 1-bit
            remove u and all edges incident on u
            from G
        else // u has 0-bit
            remove all edges and vertices
            incident on u  from G
    }
    Use Algorithm B to find vertex cover on
    tree components and remove them from G
    Remove alternating vertices on simple
    cycles from G, starting from an arbitrary
    vertex
    V' = V' - V
    return V'
}
```

Algorithm C. An algorithm that finds a vertex cover from a chromosome.

## 2.2 The Vertex Cover

The largest compatible subset problem actually has already been studied in the graph theory. A vertex cover of a graph $G=(V, E)$ is a set $V' \subseteq V$ such that if $(u, v)$ is an edge of G, then either $u \in V'$ or $v \in V'$ (or both). The vertex cover problem is to find a vertex cover that minimize $|V'|$.

The largest compatible subset problem and the vertex cover problem are actually dual. The largest compatible subset problem is to find the largest subset $U'$ to keep, such that the remaining graph is edge free, while the vertex cover problem is to find the smallest subset $V'$ that covers all edges. That is $U' = V - V'$.

The vertex cover problem has been proven to be NP-hard. Therefore, there is no efficient algorithm to solve it (and the largest compatible subset problem too). Different heuristic algorithms have been proposed to find near-optimal solution for the vertex cover problem. Algorithm A will produce a vertex cover that is at most twice the size of the optimal vertex cover. Although there is no efficient algorithm to solve the vertex cover problem in general, the optimal solution can easily be found if the graph is composed of disjointed simple cycles and trees. The optimal vertex cover for graph that is a simple cycle can be obtained by taking alternating vertices on the cycle starting from an arbitrary vertex. The optimal vertex cover for a graph that is a tree can be found by using Algorithm B.

## 3   The Proposed Method

The idea of the proposed method is to encode the vertex cover by a binary string, where each bit corresponds to a vertex in the graph. When a vertex $u$ has a 1-bit, it represents including $u$ in $V'$, and (conceptually) excluding $u$ from $V'$ otherwise. However, the optimal presence can be found for the rest of the vertices, once the presence of some "critical vertices" is determined. Therefore, our representation of the chromosome only composes of the set of critical vertices. We define the set of critical vertices to be the set $C \subseteq V$, where $\{u \in V \mid u$ has an edge degree larger than two and $u$ belongs to a cycle in $G\}$. The following lemma shows the definition of critical vertices is sufficient to decompose the graph into a graph that composes of only simple cycles or/and trees.

> **Lemma:** The set of non-critical vertices spans a graph $G'$ that is composed of only simple cycles or/and trees.
>
> Proof:
> If there exist a component in $G'$ that is not a simple cycle nor a tree, then there must exist a vertex $u$ that belongs to a cycle and has degree larger than two. But the set of non-critical vertices ($V$-$C$) is the set of vertices that have degree less than three or does not belong to any cycle in $G$, thus it contradicts the assumption.

Once the presence of the critical vertices is determined, the presence of the rest of the vertices can easily be determined by the methods discussed in Section 3. However, some critical vertices can be adjacent to each other. Thus, the effect of a vertex $u$ has a 0-bit does not immediately exclude $u$ from $V'$, but it does immediately include all adjacent vertices of $u$ from $V'$. Algorithm C shows how the vertex cover is interpreted from a chromosome.

## 4   Results

We compared the proposed method with Algorithm A on randomly generated graphs with different edge densities. The edge density is used to determine the number of edges in the graph. All implementations and testing were done on a 450MHz Intel Celeron desktop with 128MB RAM running Windows 98. We tested the performance on three different edge densities 0.3, 0.6 and 0.9 with the graph size from 50 to 250 vertices. The population size was chosen to be the number of critical vertices, and we defined the fitness to be $|V'| / |V|$ (the ratio of the vertex cover size over the graph size). The algorithm terminated when the best solution did not improve in ten generations. Furthermore, single point crossovers and mutations were used. The results were an average of ten runs. We also ran the proposed method with three different crossover rates and mutation rates, (0.3,0.8), (0.5,0.5), (0.5,0.8).

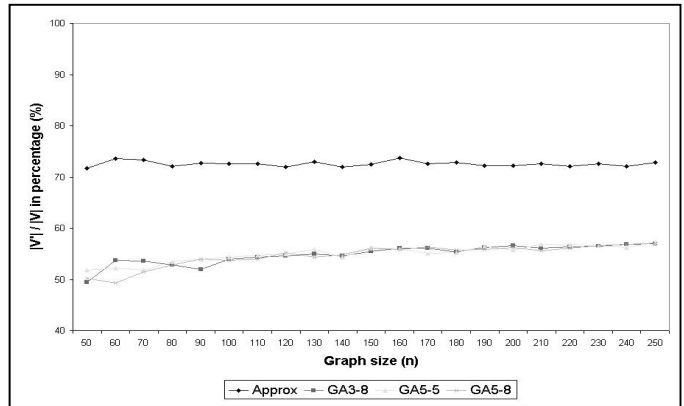

Figure 3. Performance of the proposed method versus
the 2-approximation algorithm with 0.3 edge density.

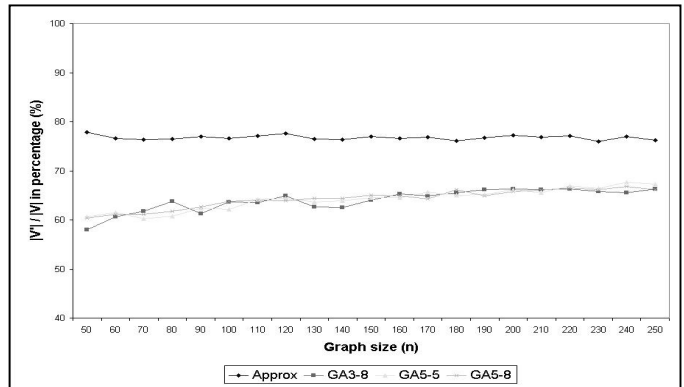

Figure 4. Performance of the proposed method versus
the 2-approximation algorithm with 0.6 edge density.

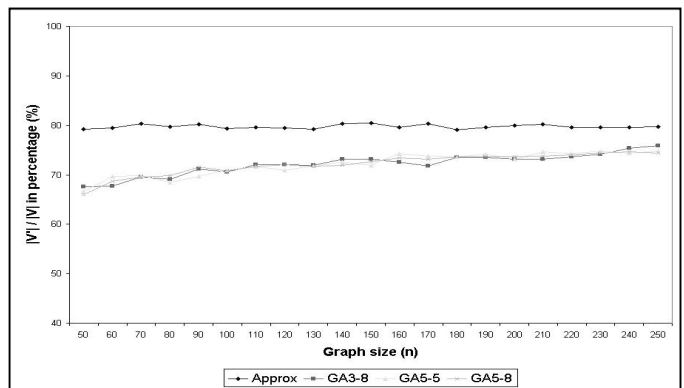

Figure 5. Performance of the proposed method versus
the 2-approximation algorithm with 0.9 edge density.

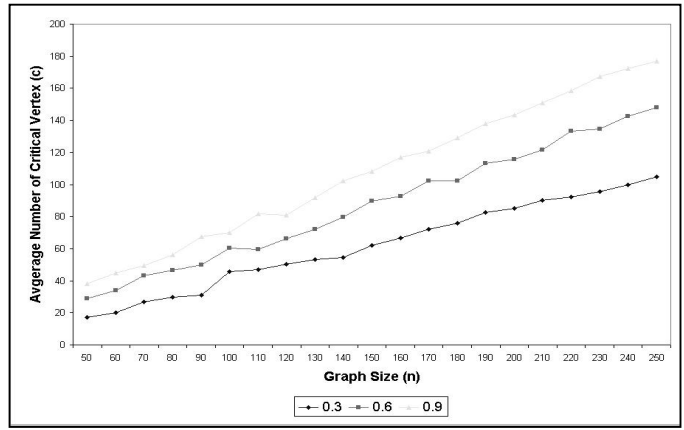

Figure 6. Number of critical vertices for various edge densities.

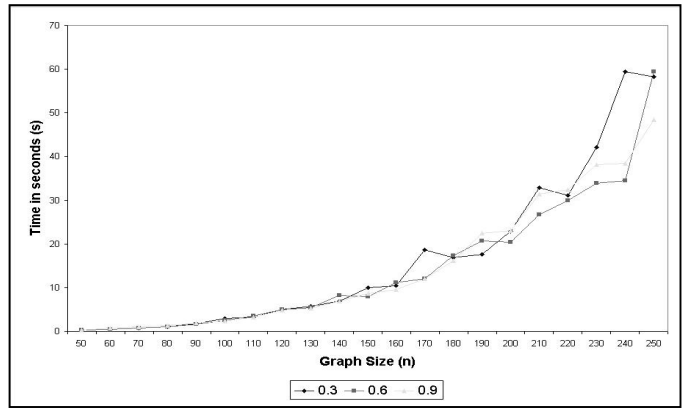

Figure 7. Computational time of the proposed method for various edge densities.

Figure 3-5 shows the ratio of the vertex cover size over the graph size (in percentage) of the proposed method (with different parameters) and of the 2-approximation algorithm for various edge densities. Due to the simple bit-wise encoding and the concept of critical vertices, the proposed method is efficient. Figure 6 shows the average number of critical vertices versus different graph size. Table 1 depicts the average

Table 1. Number of critical vertices for different edge densities 0.3, 0.6 and 0.9.

| Size | Edge density | | |
|---|---|---|---|
| | 0.3 | 0.6 | 0.9 |
| 50 | 17 | 29 | 38 |
| 100 | 45.6 | 60.2 | 70 |
| 150 | 62 | 89.6 | 108.3 |
| 200 | 85.3 | 115.9 | 143.4 |
| 250 | 104.8 | 148.2 | 177.1 |

Table 2. Computational time of the GA method for various edge densities.

| | Edge density | | |
|---|---|---|---|
| Size | 0.3 | 0.6 | 0.9 |
| 50 | 0.26 | 0.33 | 0.35 |
| 100 | 2.90 | 2.54 | 2.29 |
| 150 | 9.98 | 7.96 | 8.61 |
| 200 | 22.85 | 20.44 | 23.02 |
| 250 | 58.27 | 59.38 | 48.36 |

Table 3. Average improvement in different edge densities for various genetic parameters (crossover rate, mutation rate).

| Edge density | Genetic Parameters | | |
|---|---|---|---|
| | (0.3,0.5) | (0.5,0.5) | (0.5,0.8) |
| 0.3 | 17.8 | 17.6 | 17.6 |
| 0.6 | 12.8 | 12.3 | 12.7 |
| 0.9 | 7.8 | 7.7 | 7.3 |

number of critical vertices for graph sizes 50, 100, 150, 200 and 250 for various edge densities. The average computational time in different edge densities for various graph sizes when (0.5,0.5) is shown in Figure 7 and is illustrated in Table 2.

## 5 Discussion

We can see that the proposed method consistently outperformed the 2-approximation algorithm. When the edge density is 0.3, the average improvement from the proposed method over the 2-approximation algorithm is 17.7%, 12.6% when the edge density is 0.6, and 7.6% when 0.9. Thus, the improvement of the proposed method is inversely proportional to the edge density. We believe that is because when the graph is highly connected, most of the vertices must be discarded. On the other hand, when the edge density is small, many vertices can in fact be kept if the vertex cover is carefully chosen, and therefore a better improvement can be made.

We set the population size to be the number of critical vertices. But from the experimental results, we can see that it is in fact very small in term of the population size. Although it does demonstrate the idea where the population size is linearly proportional to the problem size, we believe that if a larger population size, such as $p$ times the number of critical vertices is used, where $p$ is between 200 and 500, a better improvement can be obtained. On the other hand, the computational time might also proportionally be increased.

Another possible improvement of our method is the interpretation of critical vertices. Other special structures can possibly be solved optimally. When considering other special structures, the trade off is on the time to detect these special structures and the time to find their optimal solutions versus the improvement of the quality of the solution. Besides, in our representation, when a vertex $u$ has a 0-bit, it does not immediately exclude $u$ from $V'$, it is simply because the critical vertices can be adjacent to each other. If possible, it will be more effective to prevent such "collision".

The perfect phylogeny is one the many phylogenetic models. Similar problems occur in other phylogenetic models. We demonstrated the effectiveness of the EC approach for the perfect phylogeny. However, we believe the EC approach is also suitable in other conflict models. For example, if the conflicts occur within a group of vertices, the problem can be seen as the vertex cover problem in a hyper-graph, where an edge can have more than two end points.

## 6 Conclusion

When inferring the phylogeny of the species, due to the noisy nature, conflicts are often found between subsets of the experimental data. The largest compatible subset problem is to filter out such noise from the experimental data by discarding the minimum amount of incompatible data. This problem is in fact equivalent to the vertex cover problem. Due to the complexity of the problem, there is no efficient algorithm to find the optimal solution, although optimal solution can be obtained for special structured graphs, such as simple cycles and trees. This paper introduced a hybrid EC approach that takes advantage of the known algorithms that produce the optimal solution on special structures. Experimental results showed that the proposed method consistently outperformed the classical 2-approximation algorithm, although it does not mathematically guarantee the quality of the solution. Due to the simple and effective encoding, the search space for the problem is dramatically reduced and therefore results in short computational time.